\documentclass[letterpaper, 10 pt, conference]{ieeeconf}  

\IEEEoverridecommandlockouts                              
\usepackage{verbatim}
\usepackage{graphics} 
\usepackage{amsmath,amssymb,stmaryrd,mathtools, amsfonts}
\usepackage{multirow}
\usepackage{hhline}
\usepackage{algorithm}
\usepackage[noend]{algpseudocode}
\PassOptionsToPackage{hyphens}{url}\usepackage{hyperref} 
\usepackage{dsfont}
\usepackage{algorithmicx}
\usepackage{multirow}
\usepackage[utf8]{inputenc}
\usepackage{import}
\usepackage{graphicx, graphics}
\usepackage{subcaption}
\usepackage{array}
\usepackage{color,xcolor}
\usepackage{soul}
\usepackage{booktabs}
\definecolor{todo_c}{rgb}{0.2,0.8,0.2} 

\usepackage{siunitx} 

\usepackage{svg}

\title{\LARGE \bf
ParkPredict+: Multimodal Intent and Motion Prediction for Vehicles in Parking Lots with CNN and Transformer
}

\author{Xu Shen, Matthew Lacayo, Nidhir Guggilla, and Francesco Borrelli%
\thanks{University of California, Berkeley, CA, USA (\{xu\_shen, 	
mattlacayo, 	
nidhir.guggilla, fborrelli\}@berkeley.edu).}%
}

\begin{document}

\maketitle
\thispagestyle{empty}
\pagestyle{empty}


\begin{abstract}
The problem of multimodal intent and trajectory prediction for human-driven vehicles in parking lots is addressed in this paper. Using models designed with CNN and Transformer networks, we extract temporal-spatial and contextual information from trajectory history and local bird's eye view (BEV) semantic images, and generate predictions about intent distribution and future trajectory sequences. Our methods outperform existing models in accuracy, while allowing an arbitrary number of modes, encoding complex multi-agent scenarios, and adapting to different parking maps. To train and evaluate our method, we present the first public 4K video dataset of human driving in parking lots with accurate annotation, high frame rate, and rich traffic scenarios.
\end{abstract}

\section{Introduction}
\label{sec:introduction}

While the rapid advancement of self-driving technology in the past decade has brought people much closer to an era with automated mobility, autonomous vehicles (AVs) still face great challenges in interacting with other road users safely and efficiently. In addition to having a robust perception system, the ability to make predictions and infer the potential future intents and trajectories of other vehicles will be essential for AVs to make optimal decisions.

Researchers have made great strides in the field of motion prediction. Physics model-based methods~\cite{lefevreSurveyMotionPrediction2014} such as the Kalman Filter and its variants leverage the dynamics of the vehicle to propagate the state of the vehicle forward for intuitive short-term predictions. Reachability study~\cite{leungInfusingReachabilitybasedSafety2020} also provides a formal way to measure the uncertainties of vehicle behavior for the control design.

When vehicles operate in complex environments, model-based approaches tend to suffer from the difficulty of accurate modeling, along with the burden of heavy computation. In contrast, deep learning methods have demonstrated great potential to incorporate various forms of information and generalize to new environments. Recurrent Neural Networks (RNNs) and Long Short-Term Memory (LSTM) networks~\cite{maTrafficPredictTrajectoryPrediction2019} are widely known for learning sequential data, and various research papers have also focused on adapting the networks to account for the multi-agent interactions ~\cite{alahiSocialLSTMHuman2016}. There is also interest in using Convolutional Neural Networks (CNN) to make predictions~\cite{djuricShorttermMotionPrediction2018} where vehicle trajectories and local environments can embedded in images efficiently.

In recent years, Transformer networks~\cite{vaswaniAttentionAllYou2017} have achieved great success in Natural Language Processing tasks. Their attention mechanism helps to keep track of global dependencies in input and output sequences regardless of the relative position, which overcomes the limitations of RNNs in learning from long temporal sequences~\cite{liEndtoendContextualPerception2020}. For trajectory prediction tasks~\cite{quintanarPredictingVehiclesTrajectories2021}, Transformer networks have been shown to outperform the LSTMs in many aspects, including accuracy~\cite{giuliariTransformerNetworksTrajectory2020} and interaction modeling~\cite{liEndtoendContextualPerception2020}.

Most of the existing work mentioned above focuses on pedestrians or vehicles driving in a road network. These environments feature simple dynamics or clear lane markings and traffic rules. However, for vehicles driving in a parking lot, we are faced with the following challenges:
\begin{enumerate}
    \item There is no strict enforcement of traffic rules regarding lane directions and boundaries.
    \item The vehicles need to perform complex parking maneuvers to drive in and out of the parking spots.
    \item There are few public datasets for the motion of vehicles in parking lots. The existing ones such as CNRPark+EXT~\cite{amato2016car} and CARPK~\cite{hsieh2017drone} are only for car detection in images and do not provide continuous trajectories in ground coordinates.
\end{enumerate}

The ParkPredict~\cite{shenParkPredictMotionIntent2020} work addresses the vehicle behavior prediction problem in parking lots by using LSTM and CNN networks. However, in~\cite{shenParkPredictMotionIntent2020}  the driving data was collected using a video game simulator to model just a single vehicle. Also, the problem formulation was restricted to a specific global map and could not be generalized. In this work, we are presenting ParkPredict+, an extensible approach that generates multimodal intent and trajectory prediction with CNN and Transformer. We also present the Dragon Lake Parking (DLP) dataset, the first human driving dataset in a parking lot environment. We offer the following contributions:
\begin{enumerate}
    \item We propose a CNN-based model to predict the probabilities of vehicle intents in parking lots. The model is agnostic to the global map and the number of intents.
    \item We propose a Transformer and CNN-based model to predict the future vehicle trajectory based on intent, image, and trajectory history. Multimodality is achieved by coupling the model with the top-k intent prediction.
    \item We release the DLP dataset and its Python toolkit for autonomous driving research in parking lots.
\end{enumerate}

The paper is organized as follows: Section~\ref{sec:formulation} formulates the multimodal intent and trajectory prediction problem, Section~\ref{sec:approach} elaborates on the model design of ParkPredict++, Section~\ref{sec:experiments} discusses the dataset, experiment setting, and results, and finally Section~\ref{sec:conclusion} concludes the paper.

\section{Problem Formulation}
\label{sec:formulation}

We aim to generate multimodal intent and trajectory predictions for vehicles driving in parking lots.

\subsection{Inputs}

We make predictions based on two types of input:

\subsubsection{Trajectory history}

Denote the current time step as $0$, the trajectory history $\mathcal{Z}_{\mathrm{hist}} = \left\{ z(t) \right\}^0_{t=-(N_{\mathrm{hist}}-1)} \in \mathbb{R}^{N_{\mathrm{hist}} \times 3}$ is the sequence of target vehicle states $z(t) = \left(x(t), y(t), \psi(t)\right) \in \mathbb{R}^3$ sampled backward in time from $0$ with horizon $N_{\mathrm{hist}}$ and step interval $\Delta t$. For convenience of notation, we denote the index set of history time steps by $\mathcal{T}_{\mathrm{hist}} = \{ -(N_{\mathrm{hist}}-1), \dots, 0\}$. 

To obtain better numerical properties and generalize the model to different maps, all states mentioned in this paper are local with respect to the vehicle body frame at $t=0$, therefore indicating that $z(0) \equiv (0, 0, 0)$.

\subsubsection{Local Contextual Information}

\begin{figure}
    \vspace{0.5em}
	\centering
	\begin{subfigure}[t]{0.42\columnwidth}
		\centering
		\includegraphics[width=\textwidth]{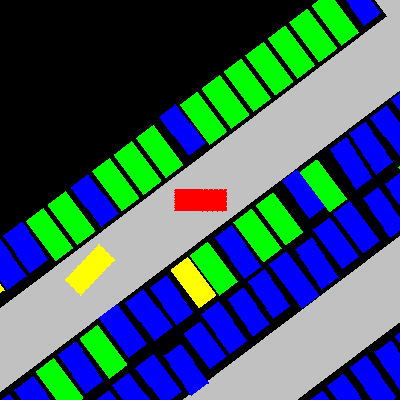}
	    \caption{\small $N_\mathrm{tail}=0$}
	    \label{fig:inst_centric_wo_tail}
	\end{subfigure}%
	~
	\begin{subfigure}[t]{0.42\columnwidth}
		\centering
		\includegraphics[width=\textwidth]{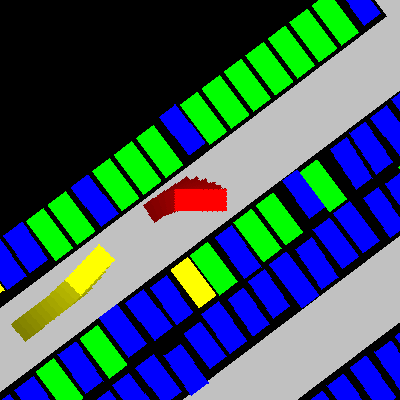}
		\caption{\small $N_\mathrm{tail}=10$.}
		\label{fig:inst_centric_10_tail}
	\end{subfigure}
	\caption{\small Rasterized BEV images with different fading tails.}
	\vspace{-1.7em}
\end{figure}

Similarly to~\cite{djuricShorttermMotionPrediction2018}, we generate a rasterized bird's eye view (BEV) image of size $n \times n$ and resolution $r$ that encodes the local environment and neighboring agents of the target vehicle. This image reflects the farthest sensing limit $L$ of the target vehicle in four directions to make decisions. Each color represents a distinct type of objects. As shown in Fig.~\ref{fig:inst_centric_wo_tail}, the driving lanes are plotted in gray, static obstacles in blue, open spaces in green, the target vehicle in red, and other agents in yellow.

Since we are interested in representing the local context of the target vehicle, we always position the target vehicle at the center pixel $(n/2, n/2)$ and rotate the image so that the vehicle always faces east. Denoting the image at time $t$ as $I(t)$, along the same horizon $\mathcal{T}_{\mathrm{hist}}$, we obtain the image histories as $\mathcal{I} = \left\{ I(t) | t \in \mathcal{T}_{\mathrm{hist}} \right\} \in \mathbb{R}^{N_{\mathrm{hist}} \times n \times n }$.

The agents' motion histories can also be encoded in a single image by plotting polygons with reduced level of brightness, resulting in a ``fading tail`` behind each agent. By setting the length of the fading tail  $N_{\mathrm{tail}}$, a longer history up to $t = - (N_\mathrm{hist} + N_{\mathrm{tail}}-1)$ can be encoded implicitly in $\mathcal{I}$. Fig.~\ref{fig:inst_centric_10_tail} shows a BEV image with $N_{\mathrm{tail}}=10$.

Note that both the trajectory and image inputs can be constructed by on-board sensors such as LiDAR. Therefore the model is adaptable to different environments without the global map or Vehicle to Everything (V2X) access.

\subsection{Outputs}

The outputs are the probability distributions over intents and future trajectory sequence:

\subsubsection{Intent}

We define the intent to be a location $\eta = (x, y)$ in the local context that the vehicle is ``aiming at`` in the long term. The vehicle does not need to reach $\eta$ at the end of the prediction horizon, but the effort to approach it would be implicitly contained along the trajectory.

\begin{figure}
    \vspace{0.5em}
    \centering
    \includegraphics[width=0.42\linewidth]{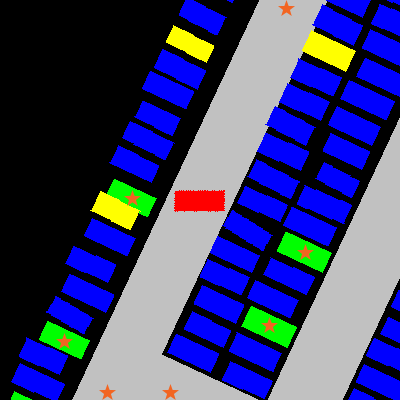}
    \caption{\small The orange stars indicate the detected possible intents, including the center of 4 empty spots and 3 lanes to drive away.}
    \label{fig:intents_example}
    \vspace{-1.2em}
\end{figure}

In this work, we are interested in two types of intents around the target vehicle: empty parking spots and lanes. We assume the existence of certain algorithms to detect all possible intents through BEV or other on-board sensor outputs, as shown in Fig.~\ref{fig:intents_example}. Given $M_{\mathrm{s}}$ empty spots and $M_{\mathrm{d}}$ lanes, the intent prediction will output a probability distribution for the $M = M_{\mathrm{s}} + M_{\mathrm{d}}$ detected intents
\begin{equation}
\label{eq:intent_prob_all}
    \hat{p} = \left[ p_{\mathrm{s}}^{[1]}, \dots, p_{\mathrm{s}}^{[M_{\mathrm{s}}]}, p_{\mathrm{d}}^{[1]}, \dots, p_{\mathrm{d}}^{[M_{\mathrm{d}}]}\right] \in \Delta^{M-1}
\end{equation}
where $p_{\mathrm{s}}^{[i]}, i\in \mathcal{N}_{\mathrm{s}} =  \{1,\dots,M_{\mathrm{s}}\}$ represents the probability that the vehicle will choose the $i$-th parking spot centered at $\eta_{\mathrm{s}}^{[i]} = (x_{\mathrm{s}}^{[i]}, y_{\mathrm{s}}^{[i]})$, and $p_{\mathrm{d}}^{[j]}, j\in \mathcal{N}_{\mathrm{d}} = \{1,\dots,M_{\mathrm{d}}\}$ represents the probability that the vehicle will bypass all spots and continue driving along the $j$-th lane towards $\eta_{\mathrm{d}}^{[j]} = (x_{\mathrm{d}}^{[j]}, y_{\mathrm{d}}^{[j]})$, which is at the boundary of its sensing limit.

\subsubsection{Future trajectories}

The trajectory output $\mathcal{Z}_{\mathrm{pred}} = \left\{ \hat{z}(t) \right\}^{N_{\mathrm{pred}}}_{t=1} \in \mathbb{R}^{N_{\mathrm{pred}} \times 3}$ is the sequence of future vehicle states along the prediction horizon $N_{\mathrm{pred}}$. 

As to be discussed in Section~\ref{sec:approach-traj}, the trajectory prediction model takes the intent as input, so given a probability $p^{[m]}$ of choosing intent $\eta^{[m]}, m \in \{1, \dots, M\}$, the predicted trajectory $\mathcal{Z}_{\mathrm{pred}}^{[m]}$ is also distributed with probability $p^{[m]}$.

\section{ParkPredict+}
\label{sec:approach}

Our model has two main components: 1) a Convolutional Neural Network (CNN) model to predict vehicle intent based on local contextual information, and 2) a CNN-Transformer model to predict future trajectory based on the trajectory history, image history, and the predicted intent.

\subsection{Intent Prediction}
\label{sec:approach-intent}

We first consider a distribution
\begin{equation*}
    \Tilde{p} = \left[ p_{\mathrm{s}}^{[1]}, \dots, p_{\mathrm{s}}^{[M_{\mathrm{s}}]}, p_{-\mathrm{s}}\right] \in \Delta^{M_{\mathrm{s}}}
\end{equation*}
where there are only the probabilities of choosing the spots ${1, \dots, M_{\mathrm{s}}}$ and bypassing all of them (denoted $p_{-\mathrm{s}}$). Compared to Eq.~\eqref{eq:intent_prob_all}, it is intuitive to have $p_{-\mathrm{s}} = \sum_{j=1}^{M_{\mathrm{d}}} p_{\mathrm{d}}^{[j]}$.

\begin{figure}
    \vspace{0.5em}
    \centering
    \includegraphics[width=0.85\linewidth]{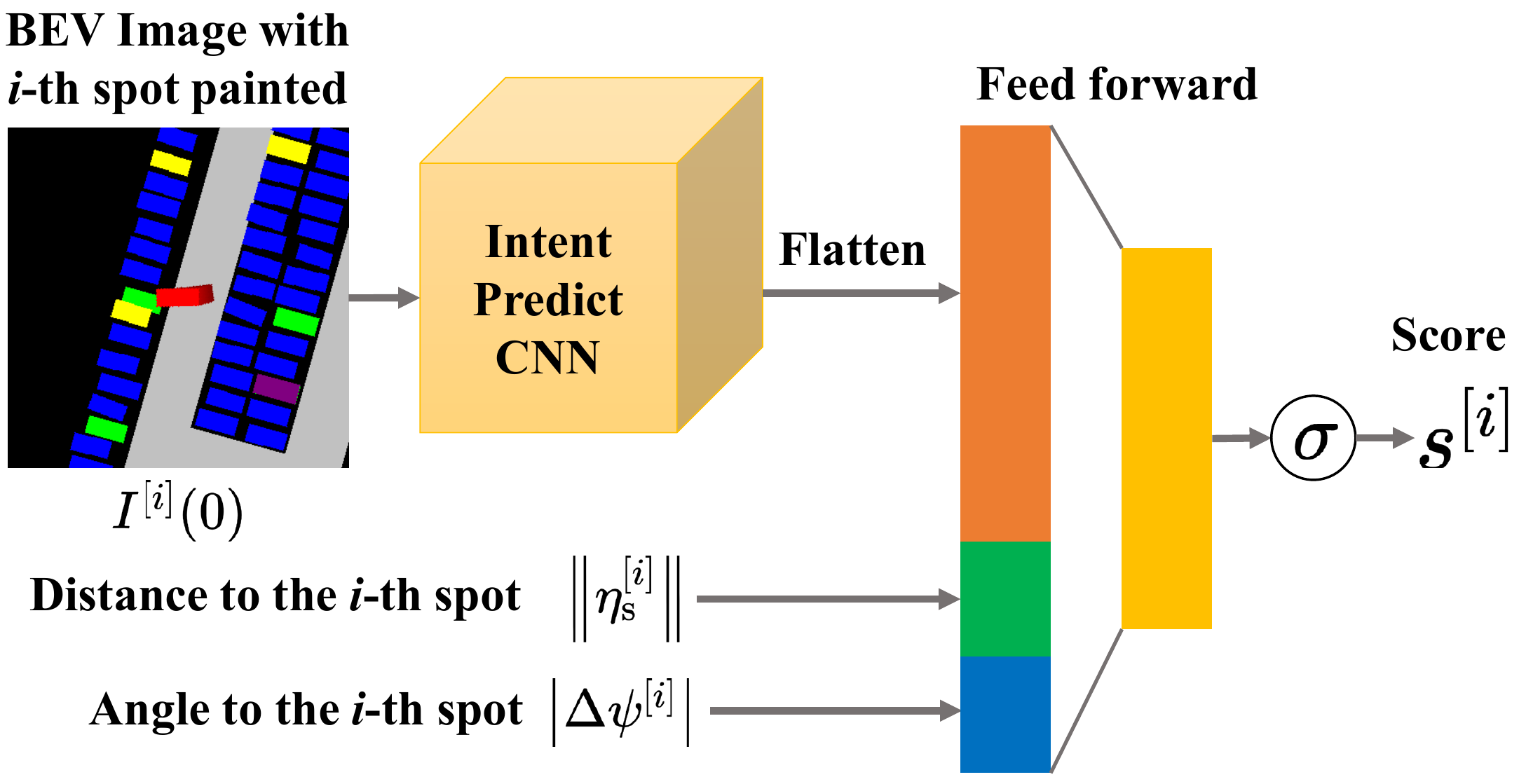}
    \caption{\small Mapping $f$ constructed by CNN and feed forward layers. The $i$-th spot is painted in purple on image $I^{[i]}(0)$.}
    \label{fig:intent_model}
    \vspace{-1.5em}
\end{figure}

Given the $i$-th spot intent $\eta_{\mathrm{s}}^{[i]} = (x_{\mathrm{s}}^{[i]}, y_{\mathrm{s}}^{[i]})$, we can generate a spot-specific image $I^{[i]}(0)$ by painting the corresponding spot a different color in the BEV image $I(0)$, as shown in Fig.~\ref{fig:intent_model}. We could also generate supplementary features for this intent, such as the distance to it $\| \eta_{\mathrm{s}}^{[i]} \|$, and the difference of heading angle $| \Delta \psi^{[i]} | = | \mathrm{arctan2} \left( y_{\mathrm{s}}^{[i]}, x_{\mathrm{s}}^{[i]} \right) |$.

Denoting by $f: \mathbb{R}^{n \times n} \times \mathbb{R} \times \mathbb{R} \rightarrow [0, 1]$ a mapping that assigns a ``score`` to an intent, we can compute the scores for all $M_{\mathrm{s}}$ spot intents as
\begin{equation}
    \hat{s}^{[i]} = f\left(I^{[i]}(0), \| \eta_{\mathrm{s}}^{[i]} \|, | \Delta \psi^{[i]} | \right), i \in \mathcal{N}_{\mathrm{s}}.
\end{equation}
Without painting any particular spot on $I(0)$ and setting the distance and angle to $0$, the score for bypassing all spots is
\begin{equation}
    \hat{s}^{[0]} = f\left(I(0), 0, 0 \right).
\end{equation}
Getting a higher score means that the corresponding intent is more likely.

We train a CNN based model as the mapping $f$. The model architecture is demonstrated in Fig.~\ref{fig:intent_model}. The loss function is binary cross-entropy (BCE) between the predicted score $\hat{s}^{[i]}$ and the ground truth label $s^{[i]}_\mathrm{gt}$
\begin{equation}
    \ell = - \left[s^{[i]}_{\mathrm{gt}} \log\hat{s}^{[i]} + \left( 1 - s^{[i]}_{\mathrm{gt}}\right) \log \left( 1-\hat{s}^{[i]} \right)\right] ,
\end{equation}
where $s^{[i]}_{\mathrm{gt}}=1$ if the vehicle chooses the $i$-th spot, and $s^{[i]}_{\mathrm{gt}}=0$ if not.

The probability outputs are normalized scores
\begin{subequations}
\begin{align}
    p_{\mathrm{s}}^{[i]} & = \frac{\hat{s}^{[i]}}{\sum_{j=0}^{M_{\mathrm{s}}} \hat{s}^{[j]}}, i \in \mathcal{N}_{\mathrm{s}}, \\
    p_{-\mathrm{s}} & = \frac{\hat{s}^{[0]}}{\sum_{j=0}^{M_{\mathrm{s}}} \hat{s}^{[j]}}.
\end{align}
\end{subequations}

To split $p_{-\mathrm{s}}$ into the probabilities of continuing to drive through different lanes, we generate a set of weights $\left\{ w^{[j]} \in [0,1] | j \in \mathcal{N}_{\mathrm{d}} \right\}$ as an arithmetic sequence and reorder them based on a simple heuristic: an intent $\eta_{\mathrm{d}}^{[j]}$ which requires more steering and longer driving distance will have lower weight $w^{[j]}$. Then, the probabilities are split as
\begin{equation}
    p_{\mathrm{d}}^{[j]} = \frac{w^{[j]}}{\sum_{k=1}^{M_{\mathrm{d}}} w^{[k]}} p_{-\mathrm{s}}, j \in \mathcal{N}_{\mathrm{d}}.
\end{equation}

We would like to highlight here that since $f$ is only a mapping from a single intent to its score, the intent prediction model proposed above is invariant to the number of detected intents $M_{\mathrm{s}}$ and $M_{\mathrm{d}}$ in the local context.

\subsection{Trajectory Prediction}
\label{sec:approach-traj}

\begin{figure}
    \vspace{0.5em}
    \centering
    \includegraphics[width=0.75\linewidth]{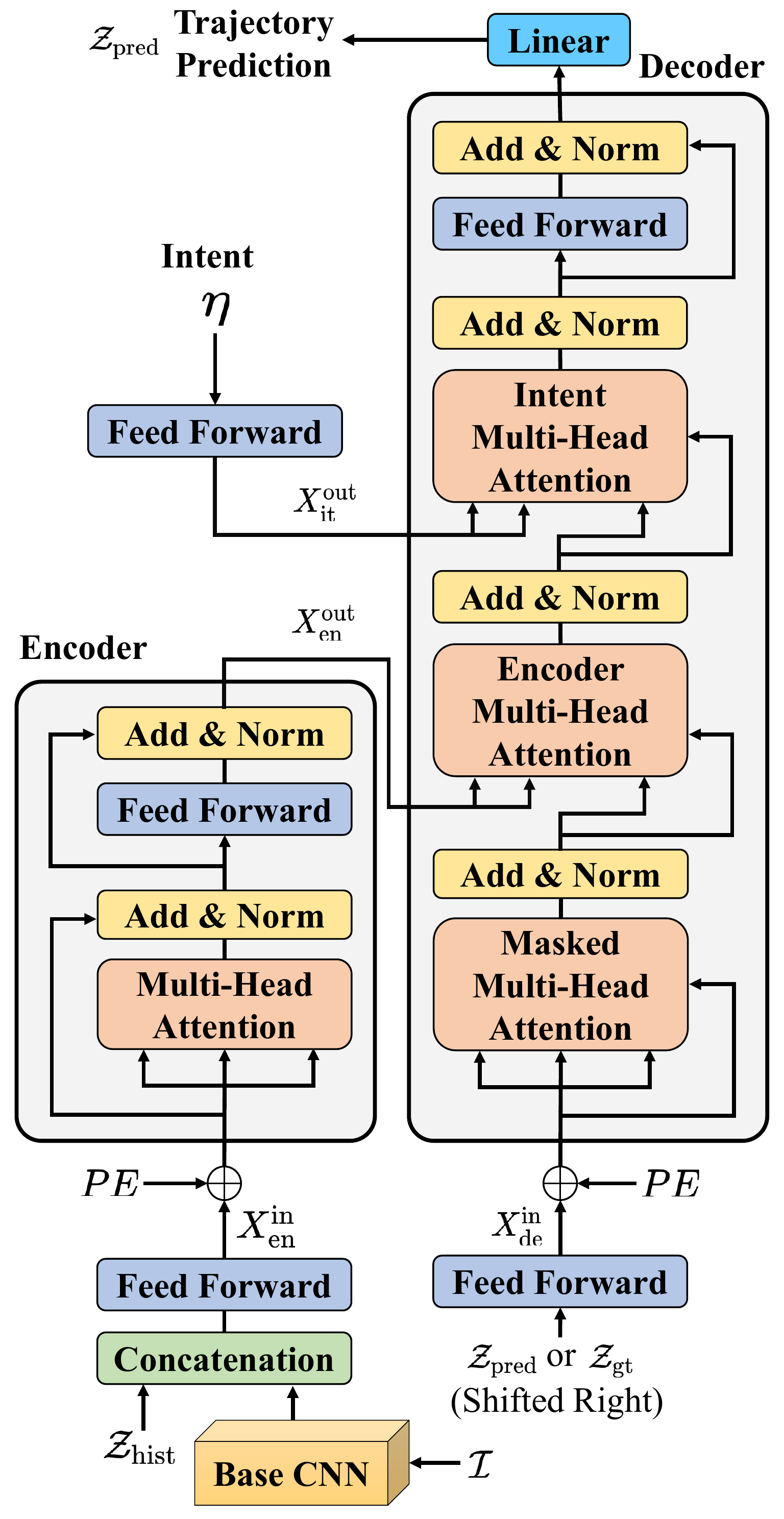}
    \caption{\small Trajectory prediction model based on Transformer.}
    \label{fig:traj_model}
    \vspace{-1.5em}
\end{figure}

We leverage the Multi-Head Attention mechanism and Transformer architecture~\cite{vaswaniAttentionAllYou2017} to construct the trajectory prediction model $\mathcal{F}: \mathbb{R}^{N_{\mathrm{hist}} \times 3} \times \mathbb{R}^{N_{\mathrm{hist}} \times n \times n} \times \mathbb{R}^{2} \rightarrow \mathbb{R}^{N_{\mathrm{pred}} \times 3}$, which predicts future trajectory $\mathcal{Z}_{\mathrm{pred}}$ with the trajectory history $\mathcal{Z}_{\mathrm{hist}}$, image history $\mathcal{I}_{\mathrm{hist}}$, and intent $\eta$. The model structure is illustrated in Fig.~\ref{fig:traj_model} and will be elaborated below.

\subsubsection{Positional Encoding}
As pointed out in~\cite{vaswaniAttentionAllYou2017}, the Attention mechanism does not contain recurrence or convolution, so we need to inject unique position encoding to inform the model about the relative position of data along the horizon. The positional encoding mask $PE$ is calculated as sinusoidal waves~\cite{suiJointIntentionTrajectory2021}:
\begin{equation}
\label{eq:positional_encoding}
    PE_{t, i} = \left\{ 
    \begin{aligned}
        & \sin \left( \frac{t}{10000^{i/D}} \right), i = 2k \\
        & \cos \left( \frac{t}{10000^{(i-1)/D}} \right), i = 2k+1
    \end{aligned}
    \right.
\end{equation}
where $t$ denotes the time step along the input horizon, $D$ denotes the model dimension, $i \in \{1, \dots, D\}$.

\subsubsection{Transformer Encoder}
\label{sec:transformer_encoder}

The image history $\mathcal{I}$ is first processed by a base CNN network $g: \mathbb{R}^{n \times n} \rightarrow \mathbb{R}^{d_{\mathrm{img}}}$ to encode contextual information: $X^{\mathrm{in}}_{\mathrm{img}}(t) = g(I(t)), t \in  \mathcal{T}_{\mathrm{hist}}$.

Subsequently the processed image history  is concatenated with the trajectory history, projected to the model dimension $D$ with a linear layer $\Phi_{\mathrm{en}}: \mathbb{R}^{d_{\mathrm{img}}+3} \rightarrow \mathbb{R}^{D}$, and summed with the positional encoding as in Eq.~\eqref{eq:positional_encoding}:
\begin{equation}
\small
    X^{\mathrm{in}}_{\mathrm{en}}(t) = \Phi_{\mathrm{en}}\left[ \mathrm{Concat}\left( X^{\mathrm{in}}_{\mathrm{img}}(t), z(t)\right) \right] \oplus PE_{t}, t \in  \mathcal{T}_{\mathrm{hist}}
\end{equation}

We then apply a  classical Transformer Encoder $\mathcal{F}_{\mathrm{en}}: \mathbb{R}^{N_{\mathrm{hist}} \times D} \rightarrow \mathbb{R}^{N_{\mathrm{hist}} \times D}$ that consists of $n_{\mathrm{head}}$ self-attention layers and a fully connected layer: $X^{\mathrm{out}}_{\mathrm{en}} = \mathcal{F}_{\mathrm{en}}(X^{\mathrm{in}}_{\mathrm{en}}).$
Residual connections are employed around each layer.
 
\subsubsection{Intent Embedding}

Given a certain intent $\eta \in \mathbb{R}^2$, we apply a fully connected layer $\Phi_{\mathrm{it}}: \mathbb{R}^2 \rightarrow \mathbb{R}^D$ to embed it as latent state for the Transformer Decoder: $X^{\mathrm{out}}_{\mathrm{it}} = \Phi_{\mathrm{it}}(\eta).$


We use the ground truth intent to train the model, and at run time we obtain multimodal trajectory predictions by picking intents with high probabilities.

\subsubsection{Transformer Decoder}

The decoder predicts the future trajectory in an autoregressive fashion. The input of the decoder $X_{\mathrm{de}}^{\mathrm{in}} \in \mathbb{R}^{N_{\mathrm{{pred}}} \times D}$ is the trajectory prediction shifted one step to the right, together with a linear projection and positional encoding as the encoder in Sect.~\ref{sec:transformer_encoder}. The Masked Multi-Head Attention block also prevents the decoder from looking ahead to future time steps.

The Encoder Multi-Head Attention block uses the Transformer Encoder output $X^{\mathrm{out}}_{\mathrm{en}} \in \mathbb{R}^{N_{\mathrm{hist}} \times D}$ as the key ($K$) and value ($V$), and the output of the Masked Multi-Head Attention block $X^{\mathrm{out}}_{\mathrm{mm}} \in \mathbb{R}^{N_{\mathrm{pred}} \times D}$ as the query ($Q$) to compute cross-attention
\begin{equation}
    \mathrm{attn}_{\mathrm{en}}=\mathrm{softmax}\left(\frac{X^{\mathrm{out}}_{\mathrm{mm}} X^{\mathrm{out}, \top}_{\mathrm{en}}}{\sqrt{D}}\right) X^{\mathrm{out}}_{\mathrm{en}}.
\end{equation}

We add the third block, Intent Multi-Head Attention, to compute the attention weights using intent so that the final trajectory prediction will be affected by intent. Here, we use intent embedding $X^{\mathrm{out}}_{\mathrm{it}} \in \mathbb{R}^{D}$ as key ($K$) and value ($V$), and the output of the previous Multi-Head Attention block $X^{\mathrm{out}}_{\mathrm{ma}} \in \mathbb{R}^{N_{\mathrm{pred}} \times D}$
\begin{equation}
    \mathrm{attn}_{\mathrm{it}}=\mathrm{softmax}\left(\frac{X^{\mathrm{out}}_{\mathrm{ma}} X^{\mathrm{out}, \top}_{\mathrm{it}}}{\sqrt{D}}\right) X^{\mathrm{out}}_{\mathrm{it}}.
\end{equation}

Finally, we apply fully connected layers at the end to generate the trajectory output $\mathcal{Z}_{\mathrm{pred}} = \left\{ \hat{z}(t) \right\}^{N_{\mathrm{pred}}}_{t=1} \in \mathbb{R}^{N_{\mathrm{pred}} \times 3}$. Residual connections are also employed in the decoder. 

The loss function is L1 loss since 1) the gradient does not decrease as the prediction gets closer to the ground truth, and 2) L1 is more resistant to outliers in the dataset:

\section{Experiments}
\label{sec:experiments}

\subsection{Dataset}
\label{sec:exp-dataset}

\begin{figure}
    \vspace{0.5em}
	\centering
	\begin{subfigure}[t]{0.48\columnwidth}
		\centering
		\includegraphics[width=\textwidth]{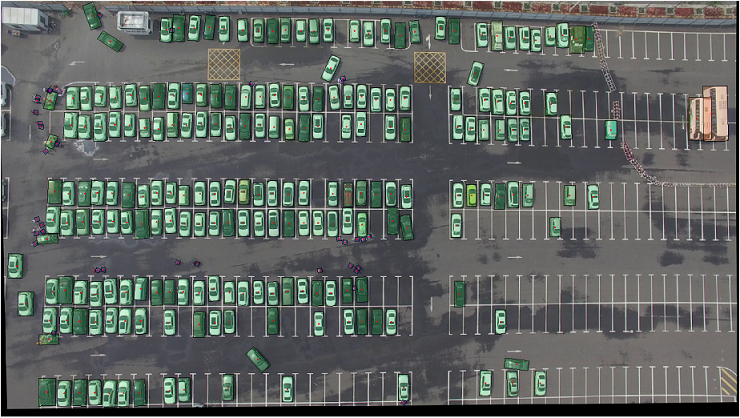}
	    \caption{\small Annotated video data.}
	    \label{fig:dlp_annotation}
	\end{subfigure}%
	~
	\begin{subfigure}[t]{0.48\columnwidth}
		\centering
		\includegraphics[width=\textwidth]{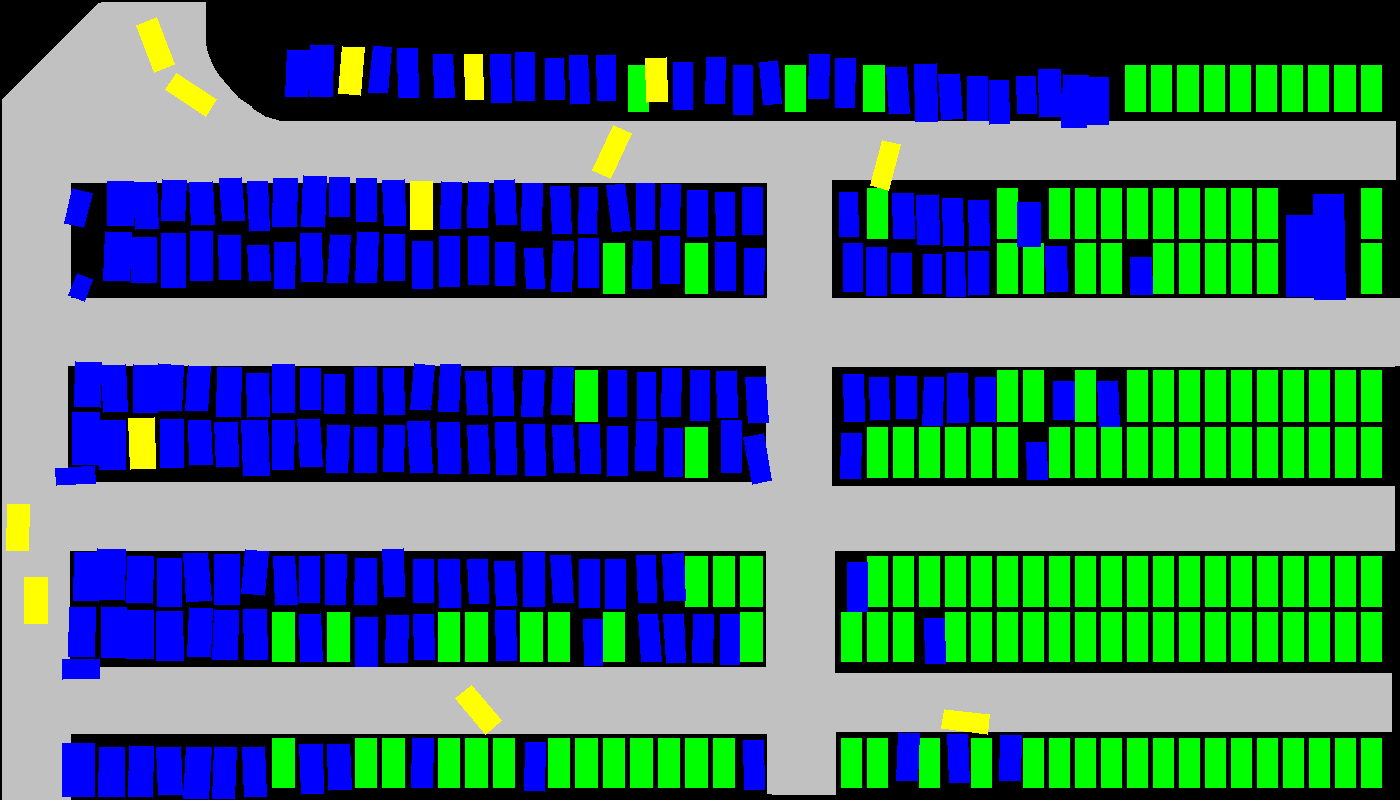}
		\caption{\small Rasterized semantic view.}
		\label{fig:dlp_semantic}
	\end{subfigure}
	\caption{\small DLP dataset.}
	\vspace{-1em}
\end{figure}

\begin{figure}
    \centering
    \includegraphics[width=\linewidth]{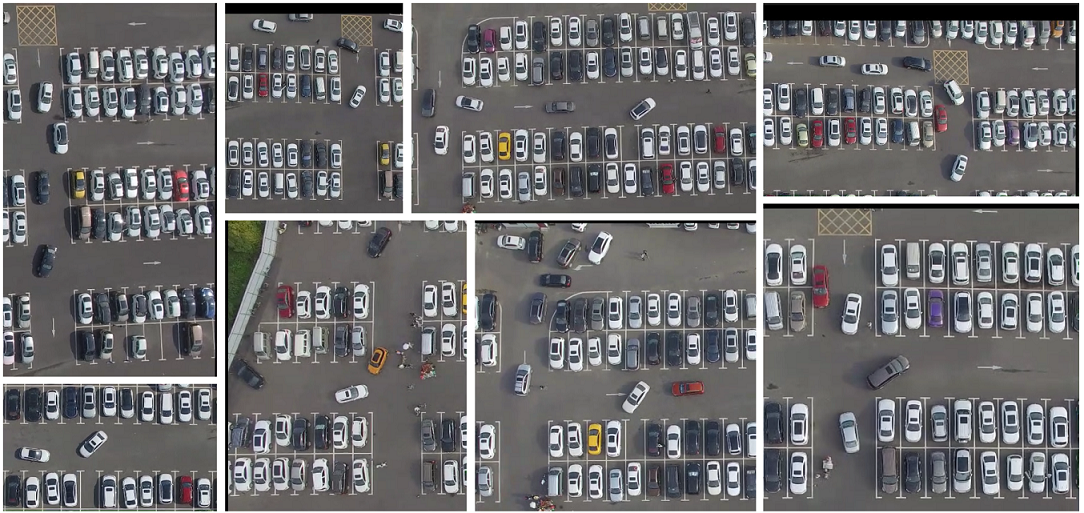}
    \caption{\small Interaction-intense scenarios in DLP dataset.}
    \label{fig:dlp_interaction}
    \vspace{-2em}
\end{figure}

We collected 3.5 hours of video data by flying a drone above a huge parking lot and named it as Dragon Lake Parking (DLP) dataset~\footnote{ \href{https://sites.google.com/berkeley.edu/dlp-dataset}{https://sites.google.com/berkeley.edu/dlp-dataset}.}. The videos were taken in 4K resolution, covering a parking area of 140 m $\times$ 80 m with about 400 parking spots (Fig.~\ref{fig:dlp_annotation}). With precise annotation, we obtain the dimension, position, heading, velocity, and acceleration of all vehicles at 25 fps. Abundant vehicle parking maneuvers and interactions are recorded, as shown in Fig.~\ref{fig:dlp_interaction}. To the best of our knowledge, this is the first and largest public dataset designated for the parking scenario, featuring high data accuracy and a rich variety of realistic human driving behavior.

We are also releasing a Python toolkit which provides convenient APIs to query and visualize data (Fig.~\ref{fig:dlp_semantic} and all BEV images included in this paper). The data is organized in a graph structure with the following components:
\begin{enumerate}
    \item Agent: An agent is an object that has moved in this scene. It contains the object's type, dimension, and trajectory as a list of instances.
    \item Instance: An instance is the state of an agent at a time step, which includes position, orientation, velocity, and acceleration. It also points to the preceding / subsequent instance along the agent's trajectory.
    \item Frame: A frame is a discrete sample from the recording. It contains a list of visible instances at this time step, and points to the preceding / subsequent frame.
    \item Obstacle: Obstacles are vehicles that never move in this recording.
    \item Scene: A scene represents a consecutive video recording with certain length. It points to all frames, agents, and obstacles in this recording.
\end{enumerate}
The entire DLP dataset contains $30$ scenes, $317,873$ frames, $5,188$ agents, and $15,383,737$ instances.

In this work, we are using the sensing limit $L=10$ m and resolution $r = 0.1$ m/pixel, so that the semantic BEV image $I$ is of size $200\times 200$. The sampling time interval $\Delta t = 0.4$s, $N_\mathrm{hist} = N_\mathrm{tail} = N_\mathrm{pred} = 10$. In other words, there are total of $8$s' information history encoded in the inputs and we are predicting the vehicle trajectory over the next $4$s.

After data cleaning, filtering, and random shuffling, we obtain a training set of size 51750 and a validation set of size 5750. The models are trained~\footnote{ { \href{https://github.com/XuShenLZ/ParkSim/}{https://github.com/XuShenLZ/ParkSim/} } .} on an Alienware Area 51m PC with 9th Gen Intel Core i9-9900K, 32GB RAM, and NVIDIA GeForce RTX 2080 GPU.

Since most well-known motion prediction benchmarks are not for the parking scenario, we choose the physics-based EKF model presented in the ParkPredict~\cite{shenParkPredictMotionIntent2020} paper as our baseline. The CNN-LSTM approach requires a different global map encoding thus cannot establish a fair comparison.

\subsection{Intent Prediction}
\label{sec:exp-intent}

\subsubsection{Hyperparameters}
\begin{table}[t]
\vspace{0.5em}
\centering
\caption{\small Hyperparameters of the Intent Prediction Model}
\label{tab:intent_parameter}
\begin{tabular}{@{}ll@{}}
\toprule
\multicolumn{1}{c}{Types}                             & \multicolumn{1}{c}{Parameters}            \\ \midrule
Conv2d $\rightarrow$ BatchNorm                        & 8 $\times$ (7 $\times$ 7) $\rightarrow$ 8            \\
Dropout $\rightarrow$ LeakyReLU $\rightarrow$ MaxPool & 0.2 $\rightarrow$ 0.01 $\rightarrow$ 2 \\ \midrule
Conv2d $\rightarrow$ BatchNorm                        & 8 $\times$ (5 $\times$ 5) $\rightarrow$ 8            \\
Dropout $\rightarrow$ LeakyReLU $\rightarrow$ MaxPool & 0.2 $\rightarrow$ 0.01 $\rightarrow$ 2 \\ \midrule
Conv2d $\rightarrow$ BatchNorm                        & 3 $\times$ (3 $\times$ 3) $\rightarrow$ 3            \\
Dropout $\rightarrow$ LeakyReLU $\rightarrow$ MaxPool & 0.2 $\rightarrow$ 0.01 $\rightarrow$ 2 \\ \midrule
Linear                                                & 6629 $\times$ 100                             \\
Linear \& Sigmoid                          & 100 $\times$ 1                               
\end{tabular}
\vspace{-1em}
\end{table}

The hyperparameters of the intent prediction model in Fig.~\ref{fig:intent_model} are outlined in Table.~\ref{tab:intent_parameter}. We use the Adam optimizer with learning rate $0.001$ and stop early at convergence.

\subsubsection{Evaluation Metrics}

We use the top-$k$ accuracy to evaluate our model: Let the set $\mathcal{N}^{(i)}_{\mathrm{k}} \subseteq \{1, \dots, M^{(i)}\}$ include the $k$ most likely intent indices in the $i$-th predicted distribution $\hat{p}^{(i)}$ and the ground truth intent be at index $l^{(i)}$, then the top-$k$ accuracy $\mathcal{A}_k$ is computed as
\begin{equation*}
    \mathcal{A}_k = \frac{1}{M_\mathrm{D}}  \sum_{i=1}^{M_\mathrm{D}} \mathbb{I} \left( l^{(i)} \in \mathcal{N}^{(i)}_{\mathrm{k}} \right).
\end{equation*}
The variable $M_\mathrm{D}$ here corresponds to the cardinality of the validation set and $\mathbb{I}(\cdot)$ is the indicator function.

\subsubsection{Results}

\begin{figure}
    \vspace{0.5em}
    \centering
    \includegraphics[width=0.85\linewidth]{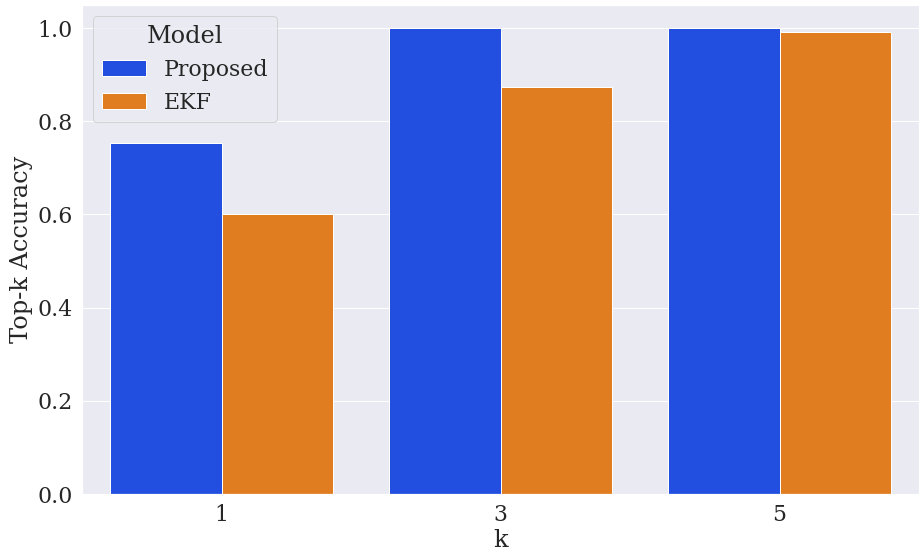}
    \caption{\small Top-k accuracy of intent prediction}
    \label{fig:intent_top_k}
    \vspace{-1.5em}
\end{figure}

Fig.~\ref{fig:intent_top_k} shows the Top-k prediction accuracy up to $k=5$. The proposed method outperforms EKF methods for all values of $k$ and achieves almost 100\% accuracy in top-3 results, which means we can reliably cover the ground truth intent when making multimodal predictions.

\subsection{Trajectory Prediction}
\label{sec:exp-traj}

\subsubsection{Hyperparameters}

We use the same Convolutional layers as Table.~\ref{tab:intent_parameter} for the Base CNN in trajectory prediction model. The subsequent feed forward layers reshape the inputs $X^{\mathrm{in}}_{\mathrm{en}}$ and $X^{\mathrm{in}}_{\mathrm{de}}$ to the model dimension $D=52$. We construct 16 encoder layers, 8 decoder layers, and 4 heads for all Multi-Head Attention blocks. The dropout rates are $0.14$ for all blocks. We use the SGD optimizer with learning rate $0.0025$ and stop early at convergence.

\subsubsection{Evaluation Metrics}

Given the $i$-th unimodal prediction $\mathcal{Z}^{(i)}_{\mathrm{pred}} = \left\{ \hat{z}^{(i)}(t) = \left( \hat{x}^{(i)}(t), \hat{y}^{(i)}(t), \hat{\psi}^{(i)}(t) \right) \right\}^{N_{\mathrm{pred}}}_{t=1}$ corresponding to the ground truth intent, we compute the mean position error $e_{\mathrm{p}}(t)$ and mean heading error $e_{\mathrm{a}}(t)$ to the ground truth label $\left\{ z^{(i)}_{\mathrm{gt}}(t) = \left( x_{\mathrm{gt}}^{(i)}(t), y_{\mathrm{gt}}^{(i)}(t), \psi_{\mathrm{gt}}^{(i)}(t) \right) \right\}^{N_{\mathrm{pred}}}_{t=1}$ as a function of the time step $t$:

\begin{subequations}
\vspace{-1em}
\begin{align*}
    e_{\mathrm{p}}(t) & = \frac{1}{M_\mathrm{D}} \sum_{i=1}^{M_\mathrm{D}} \left\|\left[\hat{x}^{(i)}(t) - x^{(i)}_{\mathrm{gt}}(t), \hat{y}^{(i)}(t) - y^{(i)}_{\mathrm{gt}}(t)\right]^{\top}\right\|_{2}, \\
    e_{\mathrm{a}}(t) & = \frac{1}{M_\mathrm{D}} \sum_{i=1}^{M_\mathrm{D}} | \hat{\psi}^{(i)}(t) - \psi^{(i)}_{\mathrm{gt}}(t) |.
\end{align*}
\end{subequations}

\subsubsection{Results}

\begin{figure}
    \vspace{0.5em}
	\centering
	\begin{subfigure}[t]{0.48\columnwidth}
		\centering
		\includegraphics[width=\textwidth]{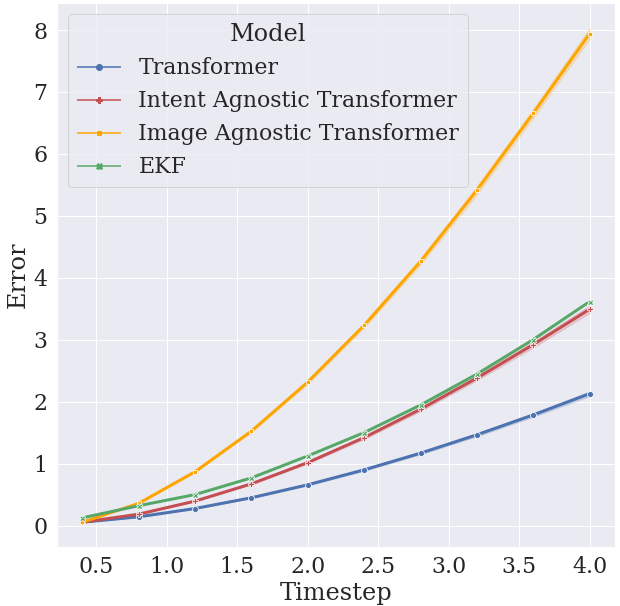}
	    \caption{\small Positional displacement.}
	    \label{fig:position_error}
	\end{subfigure}%
	~
	\begin{subfigure}[t]{0.50\columnwidth}
		\centering
		\includegraphics[width=\textwidth]{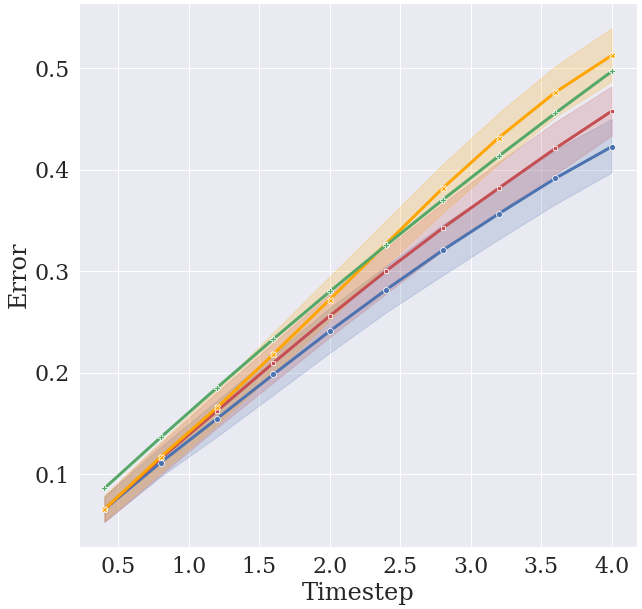}
		\caption{\small Difference of heading angle.}
		\label{fig:angular_error}
	\end{subfigure}
	\caption{\small Trajectory prediction error vs time step. Blue line represents the proposed (complete) Transformer model, red and yellow lines represent removing the intent and image inputs respectively, and green line represents the EKF model. The 95\% confidence intervals are plotted as the shaded regions.}
	\label{fig:traj_prediction_result}
	\vspace{-1.5em}
\end{figure}

We report the performance of our trajectory prediction model and the results of an ablation study in Fig.~\ref{fig:traj_prediction_result}. For positional displacement error in Fig.~\ref{fig:position_error}, we can see that our model outperform the EKF approach~\cite{shenParkPredictMotionIntent2020} in both short- and long-term. However, by removing the intent information, the model performs almost the same as the EKF, and by removing the image, the performance is severely worsened. The error of heading angle in Fig.~\ref{fig:angular_error} has less difference among models but reflects the same trend. The comparison results demonstrate that the model learns a lot of information from image and intent inputs to generate predictions with high accuracy, especially in the long-term.

\subsection{Case Study}

\begin{figure}
    \vspace{0.5em}
	\centering
	\begin{subfigure}[t]{\columnwidth}
		\centering
		\includegraphics[width=\textwidth]{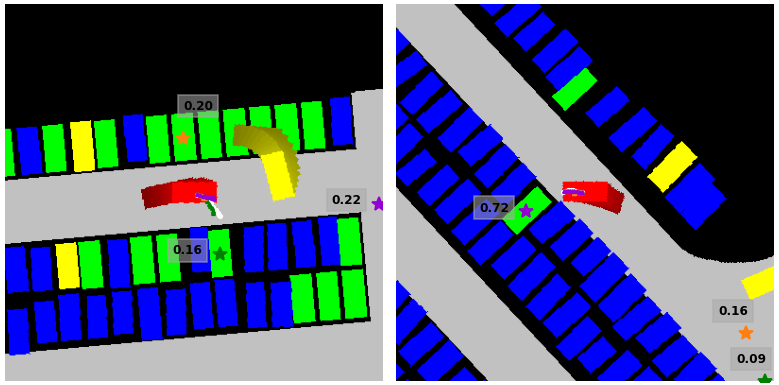}
	    \caption{\small Multi-modal prediction in the middle of parking maneuver.}
	    \label{fig:live_prediction_parking}
	\end{subfigure}%
	\\
	\begin{subfigure}[t]{\columnwidth}
		\centering
		\includegraphics[width=\textwidth]{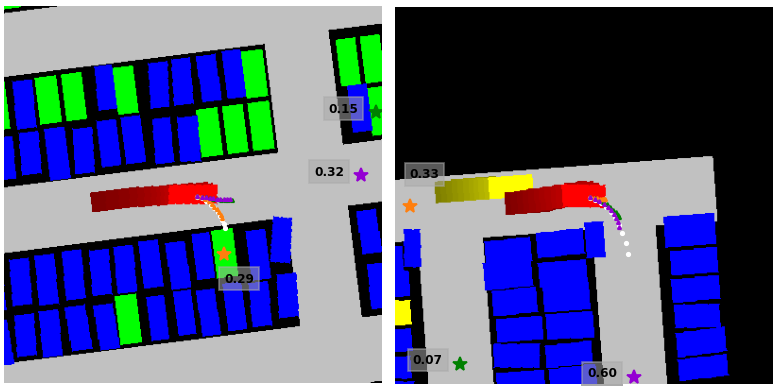}
		\caption{\small Multi-modal prediction while cruising with high speed.}
		\label{fig:live_prediction_intersection}
	\end{subfigure}
	\caption{\small Examples of multimodal prediction. The top-3 modes are visualized by orange, green, and purple. The stars and the text close by indicate the predicted intents and their probabilities. The white trajectories are the ground truth future trajectories.}
	\label{fig:live_prediction}
	\vspace{-1em}
\end{figure}

Fig.~\ref{fig:live_prediction} presents our prediction for four representative scenarios in the parking lot. The model generates accurate trajectory predictions when compared to the ground truth, while providing other possible behaviors according to the local context. Video link:~{\small \url{https://youtu.be/rjWdeuvRXp8}}.

In Fig.~\ref{fig:live_prediction_parking}, the vehicles have already slowed down and started steering towards parking spots. In the left example, it is still hard to tell the actual intention of the vehicle, therefore the top-3 intents have similar probabilities. Among them, the mode marked with green matches the ground truth to steer into the corresponding spot. At the same time, the mode in purple describes that the vehicle might continue driving towards the intersection. The mode in orange represents the case that the vehicle might prepare for a reverse parking maneuver. The trajectory prediction in orange is so short that it is mostly occluded by the other two modes, indicating the need for the car to slow down before backing up. The scenario is simpler in the right example, where the model believes with high probability that the vehicle is backing up into the spot marked by purple intent.

Fig.~\ref{fig:live_prediction_intersection} shows two other scenarios in which the vehicles are still driving with relatively high speeds (indicated by the length of the fading tail). In the left example, the model firstly predicts the most likely spot marked in orange and the resulting trajectory prediction matches the ground truth. Since there is an intersection in front, it is possible that the vehicle will continue driving across it. The model predicts this behavior with the modes in green and purple. The example on the right shows a situation where the vehicle needs to turn right at the intersection. The mode in purple successfully predicts it according to the local context. Since no empty spot is visible, the model assign some probabilities to the other two intents along the lanes, but with lower value. The orange and green trajectories also try to reach the corresponding intents by slowing the vehicle down.
\section{Conclusion}
\label{sec:conclusion}

In this work, we investigate the problem of multimodal intent and trajectory prediction for vehicles in parking lots. CNN and Transformer networks are leveraged to build the prediction model, which take the trajectory history and vehicle-centered BEV images as input. The proposed model can be flexibly applied to various types of parking maps with an arbitrary number of intents. The result shows that the model learns to predict the intent with almost 100\% accuracy in top-3 candidates, and generates multimodal trajectory rollouts. Furthermore, we release the DLP dataset with its software kit. As the first human driving dataset in parking lot, it can be applied to a wide range of prospective applications.

\section*{ACKNOWLEDGMENT}
We would like to thank Michelle Pan and Dr. Vijay Govindarajan for their contribution in building DLP dataset.

\bibliographystyle{IEEEtran}

\bibliography{ref.bib}

\begin{thebibliography}{10}
\providecommand{\url}[1]{#1}
\csname url@rmstyle\endcsname
\providecommand{\newblock}{\relax}
\providecommand{\bibinfo}[2]{#2}
\providecommand\BIBentrySTDinterwordspacing{\spaceskip=0pt\relax}
\providecommand\BIBentryALTinterwordstretchfactor{4}
\providecommand\BIBentryALTinterwordspacing{\spaceskip=\fontdimen2\font plus
\BIBentryALTinterwordstretchfactor\fontdimen3\font minus
  \fontdimen4\font\relax}
\providecommand\BIBforeignlanguage[2]{{%
\expandafter\ifx\csname l@#1\endcsname\relax
\typeout{** WARNING: IEEEtran.bst: No hyphenation pattern has been}%
\typeout{** loaded for the language `#1'. Using the pattern for}%
\typeout{** the default language instead.}%
\else
\language=\csname l@#1\endcsname
\fi
#2}}

\bibitem{lefevreSurveyMotionPrediction2014}
S.~Lef{\`e}vre, D.~Vasquez, and C.~Laugier, ``A survey on motion prediction and
  risk assessment for intelligent vehicles,'' \emph{ROBOMECH Journal}, vol.~1,
  no.~1, p.~1, Dec. 2014.

\bibitem{leungInfusingReachabilitybasedSafety2020}
K.~Leung, E.~Schmerling, M.~Zhang, M.~Chen, J.~Talbot, J.~C. Gerdes, and
  M.~Pavone, ``On infusing reachability-based safety assurance within planning
  frameworks for human\textendash robot vehicle interactions,''
  \emph{International Journal of Robotics Research}, vol.~39, no. 10-11, pp.
  1326--1345, 2020.

\bibitem{maTrafficPredictTrajectoryPrediction2019}
Y.~Ma, X.~Zhu, S.~Zhang, R.~Yang, W.~Wang, and D.~Manocha,
  ``{{TrafficPredict}}: {{Trajectory Prediction}} for {{Heterogeneous
  Traffic-Agents}},'' \emph{Proceedings of the AAAI Conference on Artificial
  Intelligence}, vol.~33, pp. 6120--6127, July 2019.

\bibitem{alahiSocialLSTMHuman2016}
A.~Alahi, K.~Goel, V.~Ramanathan, A.~Robicquet, L.~{Fei-Fei}, and S.~Savarese,
  ``Social {{LSTM}}: {{Human Trajectory Prediction}} in {{Crowded Spaces}},''
  in \emph{2016 {{IEEE Conference}} on {{Computer Vision}} and {{Pattern
  Recognition}} ({{CVPR}})}, vol.~81.\hskip 1em plus 0.5em minus 0.4em\relax
  {IEEE}, June 2016, pp. 961--971.

\bibitem{djuricShorttermMotionPrediction2018}
N.~Djuric, V.~Radosavljevic, H.~Cui, T.~Nguyen, F.-C. Chou, T.-H. Lin, and
  J.~Schneider, ``Short-term {{Motion Prediction}} of {{Traffic Actors}} for
  {{Autonomous Driving}} using {{Deep Convolutional Networks}},'' Aug. 2018.

\bibitem{vaswaniAttentionAllYou2017}
A.~Vaswani, N.~Shazeer, N.~Parmar, J.~Uszkoreit, L.~Jones, A.~N. Gomez,
  {\L}.~Kaiser, and I.~Polosukhin, ``Attention is all you need,'' in
  \emph{Proceedings of the 31st {{International Conference}} on {{Neural
  Information Processing Systems}}}, ser. {{NIPS}}'17.\hskip 1em plus 0.5em
  minus 0.4em\relax {Red Hook, NY, USA}: {Curran Associates Inc.}, Dec. 2017,
  pp. 6000--6010.

\bibitem{liEndtoendContextualPerception2020}
L.~L. Li, B.~Yang, M.~Liang, W.~Zeng, M.~Ren, S.~Segal, and R.~Urtasun,
  ``End-to-end contextual perception and prediction with interaction
  transformer,'' \emph{IEEE International Conference on Intelligent Robots and
  Systems}, pp. 5784--5791, 2020.

\bibitem{quintanarPredictingVehiclesTrajectories2021}
A.~Quintanar, D.~{Fernandez-Llorca}, I.~Parra, R.~Izquierdo, and M.~A. Sotelo,
  ``Predicting vehicles trajectories in urban scenarios with transformer
  networks and augmented information,'' \emph{IEEE Intelligent Vehicles
  Symposium, Proceedings}, vol. 2021-July, no.~Iv, pp. 1051--1056, 2021.

\bibitem{giuliariTransformerNetworksTrajectory2020}
F.~Giuliari, I.~Hasan, M.~Cristani, and F.~Galasso, ``Transformer networks for
  trajectory forecasting,'' \emph{Proceedings - International Conference on
  Pattern Recognition}, pp. 10\,335--10\,342, 2020.

\bibitem{amato2016car}
G.~Amato, F.~Carrara, F.~Falchi, C.~Gennaro, and C.~Vairo, ``Car parking
  occupancy detection using smart camera networks and deep learning,'' in
  \emph{Computers and Communication (ISCC), 2016 IEEE Symposium on}.\hskip 1em
  plus 0.5em minus 0.4em\relax IEEE, 2016, pp. 1212--1217.

\bibitem{hsieh2017drone}
M.-R. Hsieh, Y.-L. Lin, and W.~H. Hsu, ``Drone-based object counting by
  spatially regularized regional proposal network,'' in \emph{Proceedings of
  the IEEE international conference on computer vision}, 2017, pp. 4145--4153.

\bibitem{shenParkPredictMotionIntent2020}
X.~Shen, I.~Batkovic, V.~Govindarajan, P.~Falcone, T.~Darrell, and F.~Borrelli,
  ``{{ParkPredict}}: {{Motion}} and {{Intent Prediction}} of {{Vehicles}} in
  {{Parking Lots}},'' in \emph{2020 {{IEEE Intelligent Vehicles Symposium}}
  ({{IV}})}.\hskip 1em plus 0.5em minus 0.4em\relax {IEEE}, Oct. 2020, pp.
  1170--1175.

\bibitem{suiJointIntentionTrajectory2021}
Z.~Sui, Y.~Zhou, X.~Zhao, A.~Chen, and Y.~Ni, ``Joint {{Intention}} and
  {{Trajectory Prediction Based}} on {{Transformer}},'' pp. 7082--7088, 2021.

\end{thebibliography}

 

\end{document}